\documentclass{article}

\usepackage[final]{neurips_2025}

\usepackage[utf8]{inputenc} 
\usepackage[T1]{fontenc}    
\usepackage{hyperref}       
\usepackage{url}            
\usepackage{booktabs}       
\usepackage{amsfonts}       
\usepackage{nicefrac}       
\usepackage{microtype}      
\usepackage{xcolor}         
\usepackage{multirow}        
\usepackage{subcaption}
\usepackage{graphicx}
\usepackage{amsmath}
\usepackage{amssymb}
\usepackage{dsfont}
\usepackage{colortbl}
\usepackage{bm}
\usepackage{pifont}
\definecolor{Gray}{gray}{0.95}
\definecolor{Gray7}{gray}{0.75}

\usepackage{wrapfig}
\usepackage{adjustbox}
\usepackage{natbib}
\setcitestyle{numbers,square,comma}
\newcommand{\cmark}{\ding{51}}%
\newcommand{\xmark}{\ding{55}}%

\title{OGGSplat: Open Gaussian Growing for Generalizable Reconstruction with Expanded Field-of-View}

\author{
  Yanbo Wang\thanks{Equal contribution.  \textsuperscript{\textdagger}Corresponding author.} \quad
  Ziyi Wang\footnotemark[1] \quad
  Wenzhao Zheng \quad
  Jie Zhou \quad
  Jiwen Lu\textsuperscript{\textdagger} \\
  \textbf{Department of Automation, Tsinghua University, China} \\
  \footnotesize
  \texttt{\{wyb23, wziyi22\}@mails.tsinghua.edu.cn;} \\
\texttt{wenzhao.zheng@outlook.com; \{jzhou, lujiwen\}@tsinghua.edu.cn}
}

\begin{document}
\maketitle
\vspace{-10pt}
\begin{abstract}
Reconstructing semantic-aware 3D scenes from sparse views is a challenging yet essential research direction, driven by the demands of emerging applications such as virtual reality and embodied AI. Existing per-scene optimization methods require dense input views and incur high computational costs, while generalizable approaches often struggle to reconstruct regions outside the input view cone. In this paper, we propose \textbf{OGGSplat}, an open Gaussian growing method that expands the field-of-view in generalizable 3D reconstruction. Our key insight is that the semantic attributes of open Gaussians provide strong priors for image extrapolation, enabling both semantic consistency and visual plausibility. Specifically, once open Gaussians are initialized from sparse views, we introduce an RGB-semantic consistent inpainting module applied to selected rendered views. This module enforces bidirectional control between an image diffusion model and a semantic diffusion model. The inpainted regions are then lifted back into 3D space for efficient and progressive Gaussian parameter optimization. To evaluate our method, we establish a Gaussian Outpainting (GO) benchmark that assesses both semantic and generative quality of reconstructed open-vocabulary scenes. OGGSplat also demonstrates promising semantic-aware scene reconstruction capabilities when provided with two view images captured directly from a smartphone camera. Code is available at \texttt{https://github.com/Yanbo-23/OGGSplat}.
\end{abstract}

\section{Introduction}

Building realistic and semantically meaningful 3D representations of the world has become a crucial goal in computer vision, driven by applications in robotics, virtual reality, and embodied AI. Beyond reconstructing vivid textures and accurate geometry, modern systems increasingly demand semantic awareness to support high-level understanding and interaction within 3D environments. This dual demand for geometric fidelity and semantic interpretability introduces new challenges for scene representation. Recent researches typically address this by combining open-vocabulary features with 3D reconstructive representations like 3D Gaussians~\cite{kerbl20233dgs}. Approaches based on per-scene optimization~\cite{qin2024langsplat, shi2024legaussian, qu2024goi, qiu2024featuresplat, wu2024opengaussian, ye2024gaussiangrouping}, which leverage dense multi-view inputs, achieve well-structured 3D geometry with fine-grained semantic alignment. In contrast, newly emerging feed-forward methods~\cite{wang2024gsemsplat, hu2024sparselgs} offer improved scalability and generalization across scenes by predicting semantic-aware 3D representations directly from sparse input views via a trained neural network.

\begin{figure}[t]
  \centering
  \includegraphics[width=\linewidth]{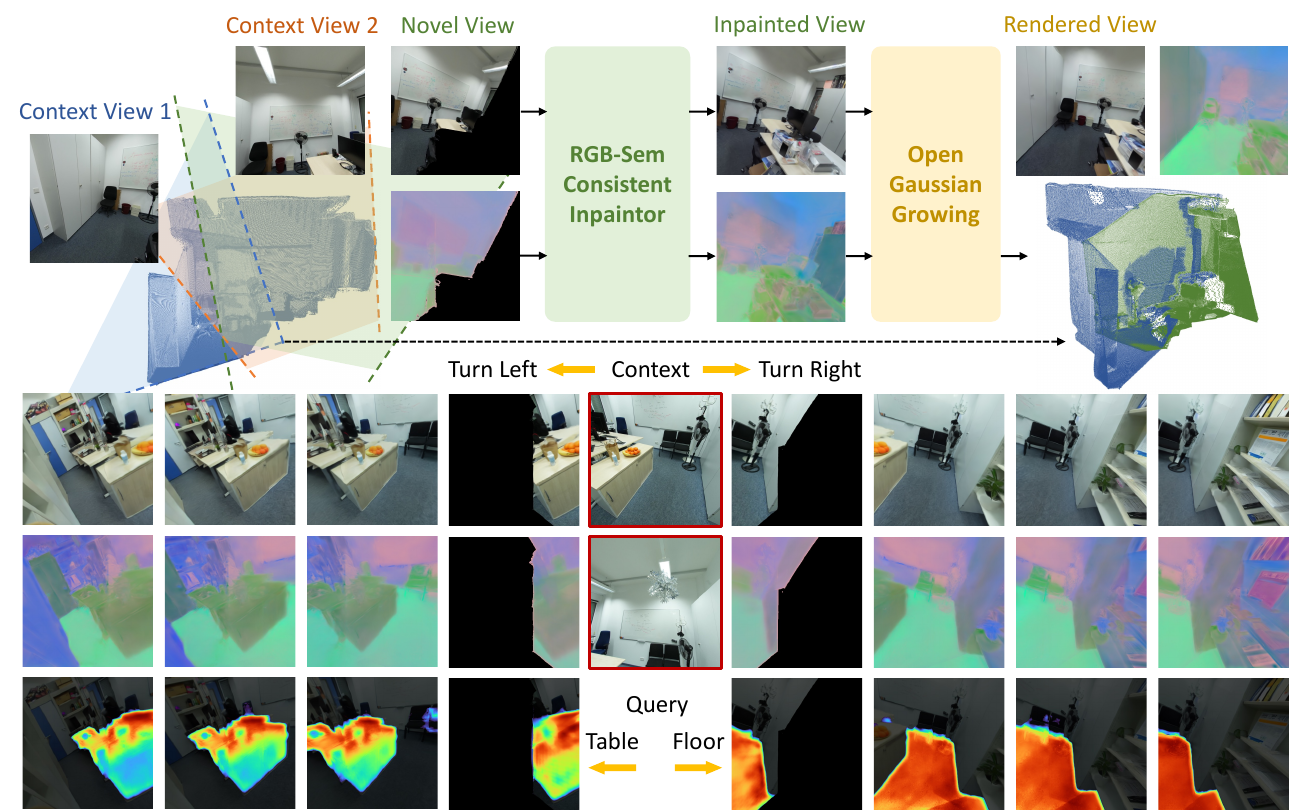}
  \vspace{-12pt}
  \caption{We propose \textbf{OGGSplat}, an open Gaussian growing method that expands the field-of-view of generalizable Gaussian reconstruction. The last three rows visualize the rendered images, their semantic maps, and category-specific heatmaps obtained by querying open-vocabulary concepts.}
  \label{fig:concept}
  \vspace{-18pt}
\end{figure}

Despite significant progress, existing methods still suffer from critical limitations. Per-scene optimization approaches typically require hundreds of input views and incur high computational time costs, often taking 25 to 50 minutes per scene. On the other hand, generalizable methods offer fast inference and handle sparse input views efficiently, but their performance is constrained by the limited scope of those inputs. When presented with extrapolated viewpoints, these models often produce distorted geometry and semantically implausible content. This highlights an urgent need for a generalizable 3D reconstruction framework that can reliably expand the field-of-view while maintaining geometric coherence and semantic consistency. We argue that incorporating semantic cues from open-vocabulary features can provide valuable guidance in imagining plausible content for unseen regions, thus extending the application of generalizable reconstruction.

In this paper, we address the aforementioned challenge of generalizable open-vocabulary 3D reconstruction by introducing \textbf{OGGSplat}, an \textbf{O}pen \textbf{G}aussian \textbf{G}rowing framework designed to extrapolate semantically meaningful 3D Gaussians beyond the input view coverage. Our goal is to enhance open-vocabulary Gaussian representations with the capacity to grow new, semantic-aware Gaussians, thereby expanding the field-of-view in scenes reconstructed from sparse inputs. A key insight of our approach is that the semantic attributes inherent in open Gaussians provide a strong prior for semantically plausible extrapolation. To exploit this, OGGSplat employs a progressive Gaussian growing strategy that builds on the initial reconstruction from sparse views. Central to this process is a novel RGB-semantic consistent inpainting module, which enables bidirectional interaction between image and semantic inpainting: semantic maps guide image completion, while inpainted images refine the semantic features in return, ensuring pixel-level alignment. The synthesized RGB images and semantic maps are then used to efficiently optimize the newly introduced Gaussians. This strategy enables OGGSplat to strike a balance between computational efficiency and the quality of the reconstructed open-vocabulary 3D scenes, even in cases of severely limited input coverage.

We conduct extensive experiments on ScanNet++~\cite{yeshwanth2023scannet++} and introduce a novel Gaussian Outpainting (GO) benchmark. Please refer to the supplementary materials for video results showcasing reconstructed scenes with expanded field-of-view and semantically coherent content. The GO benchmark is designed to assess both visual fidelity and semantic plausibility in extrapolated regions. We incorporate several state-of-the-art 2D open-vocabulary semantic segmentation models to generate the ground-truth. This enables quantitative evaluation using segmentation mean Intersection-over-Union (mIoU) in addition to commonly used generative metrics Fréchet Inception Distance (FID)~\cite{heusel2017fid}. We also deploy OGGSplat on context images captured directly using a smartphone camera. The promising results highlight its potential for future applications on portable devices.

In conclusion, the contributions can be summarized as: (1) We propose OGGSplat, the first work to expand the field-of-view for generalizable open Gaussian reconstructions. (2) We design an RGB-semantic consistent inpainting module that enforces bidirectional interaction between image and semantic map inpainting, and introduce a progressive Gaussian growing strategy to optimize new Gaussians from the inpainted content. (3) We establish the Gaussian Outpainting (GO) benchmark, enabling comprehensive evaluation with both semantic perception and generative quality metrics.

\section{Related Work}

\noindent\textbf{3D Gaussian Splatting.} 3D Gaussian Splatting (3DGS)\cite{kerbl20233dgs} is a more efficient differentiable rendering method compared with Neural Radiance Field (NeRF)~\cite{mildenhall2021nerf}. Existing 3DGS methods can be categorized by their optimization strategy and the number of input views. Early approaches~\cite{yu2024mipsplat, lu2024scaffoldgs, fan2024lightgaussian, fu2024colmapfree3dgs} rely on per-scene optimization using hundreds of images, achieving high-fidelity reconstructions at the cost of computation and scalability. Subsequent methods~\cite{xiong2023sparsegs, paliwal2024coherentgs, chung2024depthsplat, zhu2024fsgs} focus on reconstructing scenes from only a few views, though per-scene optimization is still required. More recently, generalizable methods~\cite{charatan2024pixelsplat, wang2024dust3r, smart2024splatt3r, chen2024mvsplat} emerge, which infer Gaussian parameters via a feed-forward neural network, enabling fast inference and cross-scene generalization. Building upon them, open-vocabulary 3DGS incorporates Gaussian representations with semantic features.

\noindent\textbf{Dense-view Per-scene Optimization.} The majority of open-vocabulary 3DGS methods adopt a dense-view per-scene optimization paradigm~\cite{sun2025cags, qiu2024gls, li2024instancegaussian, lu2025segmentthensplat}. LangSplat~\cite{qin2024langsplat} pioneers the field via knowledge distillation from vision-language models such as CLIP~\cite{radford2021CLIP} and DINO~\cite{caron2021DINO}. Building upon it, LEGaussians~\cite{shi2024legaussian}and GOI~\cite{qu2024goi} introduce quantization techniques to compress high-dimensional semantic embeddings into compact Gaussian parameters. Alternatively, methods such as OpenGaussian~\cite{wu2024opengaussian} and Gaussian Grouping~\cite{ye2024gaussiangrouping} utilize 2D open-vocabulary segmentation tools like SAM~\cite{kirillov2023SAM} to assign semantic labels to rendered images, without explicitly encoding semantics into the Gaussians themselves. Despite their semantic expressiveness and high-fidelity reconstructions, these approaches inherit the need for densely sampled input views and time-intensive per-scene optimization.

\noindent\textbf{Sparse-view Per-scene Optimization.} To mitigate the overfitting issue of sparse-view per-scene optimization, recent works explore view synthesis strategies. Methods such as ViewCrafter~\cite{yu2024viewcrafter} and FlowR~\cite{fischer2025flowr} use diffusion models to synthesize photometrically and geometrically consistent intermediate views. Extending this idea to open-vocabulary 3DGS, SPC-GS~\cite{liao2025spcgs} leverages video diffusion model MotionCtrl~\cite{wang2024motionctrl} to generate improved structure-from-motion initializations. To enhance semantic consistency, SPC-GS integrates SAM2~\cite{ravi2024sam2}, which provides temporally aligned semantic masks and embeddings across frames. However, incorporating video diffusion models significantly increases computation costs, and the overall optimization latency remains high.

\noindent\textbf{Generalizable Models.} Generalizable 3D reconstruction models leverage feed-forward neural networks trained on large-scale datasets to avoid per-scene optimization. PixelSplat~\cite{charatan2024pixelsplat} and MVSplat~\cite{chen2024mvsplat} rely on accurate camera pose information, while DUSt3R~\cite{wang2024dust3r} and Splatt3R~\cite{smart2024splatt3r} propose to directly infer point clouds and Gaussian parameters from unposed image pairs. The latter paradigm has quickly been extended to open-vocabulary 3DGS: GSemSPlat~\cite{wang2024gsemsplat} and SparseLGS~\cite{hu2024sparselgs} incorporate semantic prediction heads to jointly estimate open-vocabulary features alongside Gaussian parameters. Despite these advances, a key limitation is their lack of outpainting capability, where they struggle to reconstruct regions beyond the narrow visual field covered by the input views.

\section{Approach}
\label{sec:approach}
\begin{figure}[t]
  \centering
  \includegraphics[width=\linewidth]{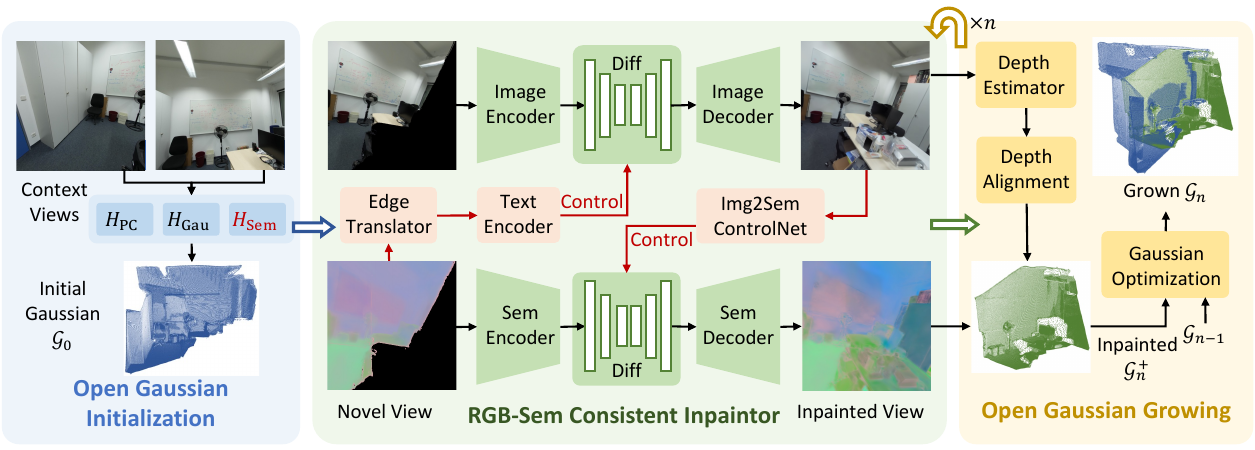}
  \vspace{-12pt}
  \caption{\textbf{OGGSplat Architecture.} We first initialize an open Gaussian reconstruction, injecting semantic features via an additional semantic head. Then, the RGB-semantic consistent inpaintor applies bidirectional controls between images and semantic maps to ensure semantic plausibility and spatial alignment. Finally, the inpainted regions are lifted back to 3D and optimized to expand the Gaussians. The last two stages are performed iteratively to progressively grow the Gaussians.}
  \label{fig:framework}
  \vspace{-16pt}
\end{figure}

As illustrated in Figure~\ref{fig:framework}, OGGSplat comprises three main stages. First, in Section~\ref{sec:semantics}, we initialize a 3D Gaussian reconstruction from the input sparse views and inject open-vocabulary semantic representations into the Gaussian parameters. Next, Section~\ref{sec:inpainting} introduces the RGB-semantic consistent inpaintor, where we propose a bidirectional control mechanism to ensure pixel-level alignment between semantics and appearance. The semantic map guides the image completion process, while the inpainted image, in turn, refines the semantic features. Finally, to allow the 3D Gaussian structure to grow consistently with the generated content, we design a progressive Gaussian growing strategy, detailed in Section~\ref{sec:growing}. The second and third stages are applied iteratively to gradually expand the Gaussian representation beyond the initial field-of-view. In practical usage, OGGSplat takes as input any two uncalibrated images and processes them through the above three stages to produce an expanded 3D Gaussian scene enriched with open-vocabulary semantics. This enables real-time rendering of both RGB images and their corresponding semantic feature maps from arbitrary viewpoints, supporting a variety of downstream tasks such as grounding and scene editing.

\subsection{Generalizable Open Gaussian Initialization}
\label{sec:semantics}

\noindent\textbf{Gaussian Reconstruction.} Given any two uncalibrated but overlapping images $I_1, I_2 \in \mathbb{R}^{H \times W \times 3}$ with height $H$ and width $W$, we adopt Splatt3R~\cite{smart2024splatt3r} to reconstruct an initial Gaussian $\mathcal{G}_0 \in \mathbb{R}^{N\times d}$ via a shared backbone, cross-attention interactions and multiple Gaussian heads. The number of Gaussian primitives $N=2\times H\times W$ corresponds to the total number of image pixels, while each Gaussian feature of dimension $d$ is composed of the following components: (1) a 3D point position $p \in \mathbb{R}^3$, (2) a position offset $p_\Delta \in \mathbb{R}^3$, defining the Gaussian center $\mu = p + p_\Delta$, (3) a rotation quaternion $q \in \mathbb{R}^4$ and a scale vector $s \in \mathbb{R}^3$, together determining the covariance matrix $\boldsymbol{\Sigma}$, (4) an opacity scalar $\alpha \in \mathbb{R}$, controlling the transparency of the Gaussian, and (5) a view-dependent appearance embedding represented by spherical harmonics $\mathbf{S} \in \mathbb{R}^{3 \times d_\mathrm{color}}$ of $d_\mathrm{color}$ degrees.

\noindent\textbf{Open Feature Injection.} To incorporate open-vocabulary clues, we introduce an additional semantic head \( H_{\text{sem}} \) to predict semantic parameters $f\in \mathbb{R}^{d_\mathrm{sem}}$ for each Gaussian primitive, inspired by GSemSplat~\cite{wang2024gsemsplat}. Following common practice~\cite{ye2024gaussiangrouping, wang2024gsemsplat}, we set the semantic embedding dimension $d_\mathrm{sem}=16$ to reduce the computational overhead during Gaussian rendering. To supervise the predicted semantic features $f$, we adopt the well-optimized vision-language APE~\cite{shen2024ape} model to efficiently obtain pixel-dense open-vocabulary semantic supervision signals $F^\mathrm{gt}\in \mathbb{R}^{H\times W\times d_\mathrm{APE}}$, where the APE semantic feature dimension $d_\mathrm{APE} \gg d_\mathrm{sem}$. To align the dimensionality, we train an autoencoder composed of a down-projection encoder $\mathcal{E}_\downarrow$ that maps the APE features to $d_\mathrm{sem}$, and a corresponding decoder $\mathcal{D}_\uparrow$ that reconstructs the original features with minimal information loss. The semantic learning objective in this stage is formulated as a cosine similarity loss:
\begin{equation}
    \mathcal{L}_{\text{sem}} = \sum_v \sum_{h, w} \left(1 - \cos\left(f_{v,h,w},\mathcal{E}_\downarrow\left(f^{gt}_{v,h,w}\right)\right)\right),
\end{equation}
where $h\in[0,H), w\in[0,W)$ denote pixel coordinates and $v$ represents target view index. The semantic feature $f_{v,h,w}$ is computed with \(\alpha\)-blending, analogous to that used for RGB rendering.

\subsection{RGB-Semantic Consistent Inpaintor}
\label{sec:inpainting}
Once the the initial Gaussian $\mathcal{G}_0$ is reconstructed, we render RGB images $I_v$ and their corresponding semantic maps $F_v$ from novel viewpoints $v$. However, when rendering outside the vision cone of the context views, hollow regions often appear due to out-of-view areas and occlusion variations, as illustrated in Figure.~\ref{fig:concept}. While pre-trained inpainting diffusion models~\cite{rombach2022ldm, lugmayr2022repaint, xie2023smartbrush} can partially address this issue, maintaining pixel-wise consistency between inpainted images and their semantic maps remains challenging. This spatial misalignment will be inherited by the following Gaussian growing process and can lead to increasingly severe semantic inconsistencies as the scene expands. Fortunately, we observe that although the semantic modality introduces challenges, it also offers valuable guidance: the partial semantic information, especially around the boundaries of incomplete regions, can be translated into explicit textual prompts to guide image inpainting. Symmetrically, inpainted RGB images can provide pixel-wise appearance cues to control semantic map completion. Therefore, we propose bidirectional controls between the RGB branch $\mathrm{Diff}_\mathrm{rgb}$ and the semantic branch $\mathrm{Diff}_\mathrm{sem}$, allowing them to mutually enhance each other during the inpainting process.

\noindent\textbf{Semantic-to-RGB Control.} To define the inpainting mask that determines whether a pixel should be inpainted, we rely on the rendered opacity $\alpha$ of each pixel. Similar to color rendering, we render an opacity map $A$, and then derive the inpainting mask $M_v$ for each view $v$ by applying a pre-defined threshold $\tau$. For simplicity, we omit the view subscript $v$ in the following discussion.
\begin{equation}
    A_{h,w} =\sum_{i\in\Theta_{h,w}}\alpha_i\prod_{j=1}^{i-1}(1-\alpha_j), \quad\quad M_{h,w} = \mathds{1}\left[ A_{h,w} < \tau \right],
\end{equation}
where $\Theta_{h,w}$ denotes the set of Gaussians contributing to the pixel at coordinate $(h,w)$.

Then we design an \textit{Edge Translator} to extract semantic concepts near the inpainting boundaries defined by the mask $M$, providing clearer guidance for filling in the hollow regions. Specifically, we first identify pixels along the boundary as $\Omega_\mathrm{edge}$. The corresponding semantic features $f_\textrm{edge}$ of these boundary pixels are then decoded into a higher-dimensional space using our pre-trained decoder $\mathcal{D}_\uparrow$:
\begin{equation}
    g_\mathrm{edge} = \mathcal{D}_\uparrow(f_\mathrm{edge}),\ \mathrm{for\ pixels\ in\ }\Omega_\mathrm{edge}
\end{equation}
Simultaneously, we prepare a set of candidate classes $\mathcal{C}_\mathrm{cand}$, consisting of the top 100 semantic categories in our training dataset. These categories are encoded into the same feature space as $g_\mathrm{edge}$. We then compute the cosine similarity between  $g_\mathrm{edge}$ and $g_\mathrm{cand}$ to perform pixel-wise segmentation:
\begin{equation}
    c_\mathrm{edge} = \mathrm{argmax}_{c_i \in \mathcal{C}_\mathrm{cand}} \mathrm{cos}(g_\mathrm{edge}, g_{c_i}),
\end{equation}
In this way, we can obtain a set of semantic categories $\mathcal{C}_\mathrm{edge}$ that are most relevant to the inpainting region. Based on these categories, we generate a prompt text $T$ in the format of \emph{``a room with $\mathrm{cate}_1$, $\mathrm{cate}_2$, ..., and $\mathrm{cate}_i$''}, which is used to guide the diffusion-based RGB image inpainting model:
\begin{equation}
    I^\mathrm{inp} =\mathrm{Diff}_{\mathrm{rgb}}(I,M,T),
\end{equation}

\noindent\textbf{RGB-to-Semantic Control.} Inspired by ControlNet~\cite{zhang2023controlnet}, we also design an RGB-to-Semantic control module to ensure that the generated semantic content aligns well with the corresponding regions in the RGB image. Formally, the completed semantic map is computed as:
\begin{equation}
    F^\mathrm{inp} = \mathrm{Diff}_{\mathrm{sem}}(F, M, T, \mathrm{ControlNet}(I^\mathrm{inp})),
\end{equation}
where $F$ is the incomplete rendered semantic feature map, and \(\mathrm{ControlNet}(I^\mathrm{inp})\) denotes the control module conditioned on the inpainted image $I^\mathrm{inp}$. Please refer to the ControlNet paper or our supplementary for further details. This module guides the semantic generation process, ensuring both structural and appearance consistency between the predicted semantic features and the RGB content.

\subsection{Open Gaussian Growing}
\label{sec:growing}

Obtaining the inpainted RGB images and semantic feature maps from selected views is not the final step of our pipeline. These results must be aggregated back into the initial Gaussian $\mathcal{G}_0$ to enable real-time rendering from arbitrary novel viewpoints. For a set of selected anchor views $V=\{v_3,v_4,\cdots,v_a\}$, we perform iterative inpainting and progressively incorporate the newly completed regions into the Gaussian. At each iteration $n$, a new view is rendered based on the currently aggregated Gaussians $\mathcal{G}_{n-1}$ and the newly inpainted content $\mathcal{G}^+_{n}$ is fused into this representation. Below, we break down a single iteration and describe the Gaussian growing process in detail.

The inpainted image $I^\mathrm{inp}$ and semantic map $F^\mathrm{inp}$ will serve as supervision targets for the newly grown Gaussians. However, establishing 3D geometry from a single novel view is inherently ill-posed, especially in regions that are newly generated during inpainting. To enrich these views with structural knowledge, we adopt custom depth estimation model~\cite{piccinelli2024unidepth, yang2024depthanything, yang2024depthanythingv2} to predict an absolute depth map $D^\mathrm{inp}$ from $I^\mathrm{inp}$. This depth map is then used to lift pixels back into 3D space, forming a point cloud in the global coordinate system. The resulting 3D points are used to initialize the position of the incremental Gaussian set $\mathcal{G}^+$, which is progressively integrated into the scene representation.
\begin{equation}
    P^+ = \operatorname{proj}(D^\mathrm{inp}, v^\mathrm{inp}, v_1, K) \cdot \beta,\ \mathrm{where} \ \beta = \frac{\sqrt{\frac{1}{M} \sum_{i=1}^{M} \left\| p^\mathrm{ori}_i \right\|_2^2}}{\sqrt{\frac{1}{N} \sum_{i=1}^{N} \left\| p^\mathrm{new}_i \right\|_2^2}}
\end{equation}
where $v^\mathrm{inp}$ and $v_1$ are the camera poses corresponding to the images $I^\mathrm{inp}$ and $I_1$, respectively, and $K$ denotes the intrinsic camera parameters. The scale factor $\beta$ is introduced to align the newly projected point cloud with the original 3D space in terms of depth. $p^\mathrm{ori},p^\mathrm{new}$ denote the original and newly projected 3D points within the overlapping regions, while \(M\) and \(N\) represent the respective number of points in each set. It is worth noticing that scaling point coordinates alone does not ensure perfect alignment. Nonetheless, it offers an efficient and approximate initialization, since the entire scene is constructed with respect to the normalized coordinate system of the first view.

At the $n^{th}$ iteration, after merging $\mathcal{G}_{n-1}$ with the newly initialized Gaussians $\mathcal{G}_n^+$, we perform efficient per-scene optimization to update the grown Gaussian $\mathcal{G}_n$. This optimization is supervised by the original sparse context views, previously and newly inpainted views. The objective function is:
\begin{align}
    \mathcal{L} &= \lambda_{\text{rgb}} \cdot \mathcal{L}_{\text{rgb}} + \lambda_{\text{feat}} \cdot \mathcal{L}_{\text{feat}}, \\
    \mathrm{where}\ \mathcal{L}_{\text{rgb}} = \lambda_1 \cdot \mathcal{L}_{\text{L1}}(I^\mathrm{r}, I^\mathrm{inp}) &+ \lambda_2 \cdot \mathcal{L}_{\text{SSIM}}(I^\mathrm{r}, I^\mathrm{inp}),\ \mathrm{and}\ \mathcal{L}_{\mathrm{feat}} =1-\mathrm{cos}(F^\mathrm{r}, F^\mathrm{inp})
\end{align}
where \(\lambda_1\) and \(\lambda_2\) balance pixel-wise accuracy and perceptual similarity, while \(\lambda_{\text{rgb}}\) and \(\lambda_{\text{feat}}\) control the overall contributions of the photometric and semantic losses, respectively. $I^\mathrm{r},F^\mathrm{r}$ denote the rendered RGB images and semantic features from the optimizing Gaussian from $v^\mathrm{inp}$.

\section{Experiments}

\subsection{The Gaussian Outpainting (GO) Benchmark}
\label{sec:benchmark}
To effectively evaluate both the visual fidelity and semantic plausibility of OGGSplat in extrapolated regions, we introduce a novel Gaussian Outpainting benchmark based on the validation set of the ScanNet++~\cite{yeshwanth2023scannet++} dataset. Detailed information about this dataset can be found in the supplementary.

\noindent\textbf{Data Composition.} 
The GO benchmark covers all 50 validation scenes from ScanNet++. For each scene, we select \textit{1 image pair} as the context views to serve as model inputs. To ensure consistency in data sampling and maintain temporal coherence, the context views are chosen as the $1^\text{st}$ and $10^\text{th}$ frames. This selection introduces moderate viewpoint variation while preserving semantic continuity, enabling a more meaningful evaluation of extrapolated content. For evaluation, we uniformly sample \textit{16 novel camera poses} within a horizontal range of $[-60^\circ, 60^\circ]$ and a vertical range of $[-20^\circ, 20^\circ]$ around the pose of the context image $I_1$. Novel RGB images and semantic maps are directly rendered from the reconstructed Gaussians at these poses and used as evaluation samples. To account for randomness in generation, we repeat the experiment five times and report the average results.

\noindent\textbf{Visual Fidelity Evaluation.} We adopt the Fréchet Inception Distance (FID)~\cite{heusel2017fid} to evaluate the statistical similarity between rendered and real images. For FID computation, all images from the validation split of the ScanNet++ dataset are used as the reference distribution. FID is then calculated between this reference distribution and the distribution of the newly rendered images. However, we observe that the limited number of generated images can negatively affect the stability of the FID metric. To address this, we increase the context views from \textit{one pair} to \textit{ten pairs} per scene, while maintaining a frame interval of 10 within each pair. This expands the number of newly rendered images by a factor of ten, resulting in a more stable and reliable FID evaluation.

\noindent\textbf{Semantic Plausibility Evaluation.} While visual fidelity is evaluated over the entire rendered image, the semantic plausibility focuses on newly outpainted regions using the mean Intersection over Union (mIoU) metric. To this end, we restrict semantic evaluation to regions rendered by the initial Gaussian that exhibit low confidence, defined as having an accumulated opacity below 0.3 in novel views. This targeted evaluation ensures that the benchmark focuses on semantic consistency in extrapolated areas. Since ground truth semantic annotations are unavailable for these extrapolated regions, we generate ground truth labels using five state-of-the-art open-vocabulary 2D semantic segmentation models~\cite{xu2023odise, shen2024ape, zeng2024maskclip++, yu2023fcclip, jiao2024maft+}. Their predictions are aggregated via a majority voting scheme, where each pixel is assigned the label most frequently predicted across the five models. To assess the quality of semantic segmentation, we follow the protocol in~\cite{kerr2023lerf, shi2024legaussian, qin2024langsplat} by computing a relevancy score for each text query. More details on relevancy score computation are provided in the supplementary. To ensure generality, we retain only those predicted mask regions with a relevancy score exceeding 50\% as the final binary mask. This filtering mechanism makes our evaluation suitable even for images where a specific category may be absent. During evaluation, we focus on 10 commonly used semantic categories selected from the top 20 classes in ScanNet++, such as \textit{wall}, \textit{floor}, \textit{chair}, \textit{table}, and others.

\label{others}
\begin{table}
  \caption{\textbf{Gaussian Outpainting (GO) benchmark results.} We compare generative metric FID and semantic metric mIoU (\%) between OGGSplat and previous methods.}
  \vspace{4pt}
  \label{tab:miou-comparison}
  \centering
  \setlength\tabcolsep{3pt}
  \resizebox{\linewidth}{!}{
  \begin{tabular}{lcc|cccccccccc}
    \toprule

    \multirow{2}{*}{Methods} & \multicolumn{1}{c}{Generation} & \multicolumn{11}{c}{Segmentation (IoU$\uparrow$)} \\
    \cmidrule(lr){2-2}\cmidrule(lr){3-13}
     & FID$\downarrow$ & \cellcolor{Gray}mIoU  
     & wall & ceiling  & floor  & table & door & (s)cabinet & chair & (b)shelf & box  & bed \\
    \midrule
    LangSplat~\cite{qin2024langsplat} & 50.4 & \cellcolor{Gray}6.9 & 29.0 & \textbf{13.4} & 15.8 & 1.8 & 4.0 & 1.3 & 2.5 &0.0 & 0.8 & 0.0  \\
    Splatt3R~\cite{smart2024splatt3r}  & 46.4 & \cellcolor{Gray}6.0  &10.1 & 2.1 & 18.9 & 5.1 & 0.0  & 1.6 & 13.8 & 0.3 &0.0  & 2.3   \\
    \textbf{OGGSplat} (Ours) & \textbf{37.5} &  \cellcolor{Gray}\textbf{17.6}  & \textbf{45.6} & 0.1 & \textbf{58.3} & \textbf{13.3} & \textbf{5.4} & \textbf{3.7} & \textbf{21.4} & \textbf{7.4} & \textbf{3.1} & \textbf{18.0} \\
    \bottomrule
  
  \end{tabular}
  }
  \vspace{-12pt}
\label{tab:go_benchmark_results}
\end{table}

\begin{figure*}[t]
  \centering
  \includegraphics[width=\linewidth]{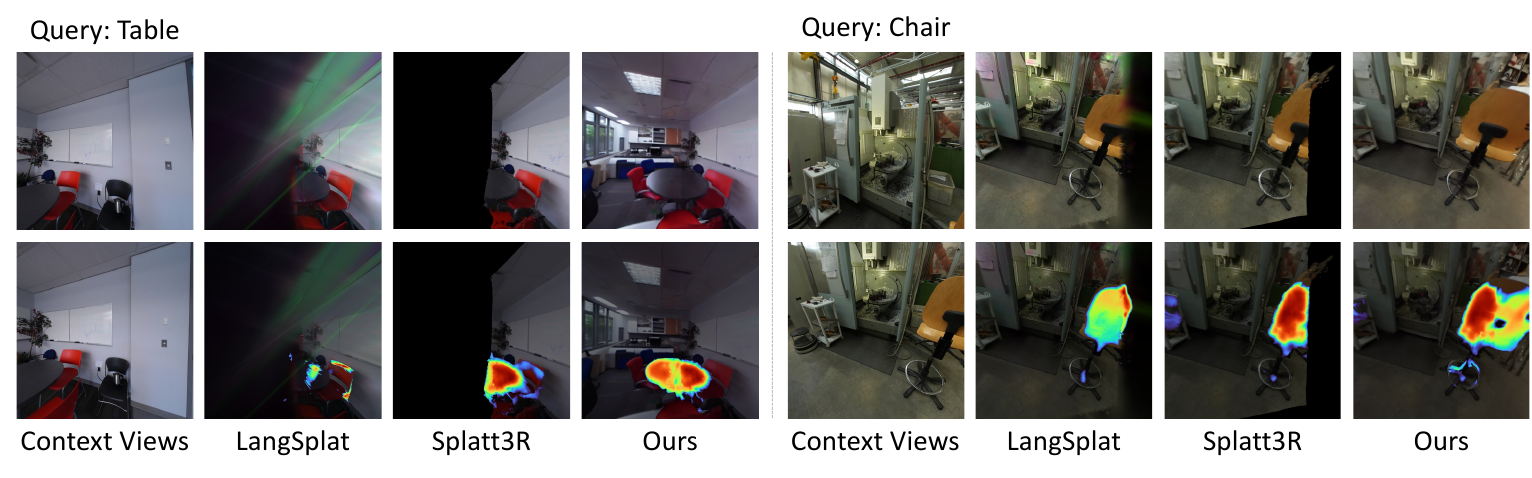}
  \vspace{-18pt}
  \caption{
    \textbf{Qualitative comparisons between LangSplat, Splatt3R, and OGGSplat on the GO benchmark.} The first row presents RGB images rendered from novel, out-of-scope viewpoints. The second row visualizes the heatmap when querying different text concepts.
  }
  \label{fig:qualitative_results}
  \vspace{-12pt}
\end{figure*}

\begin{figure*}[t]
  \centering
  \includegraphics[width=\linewidth]{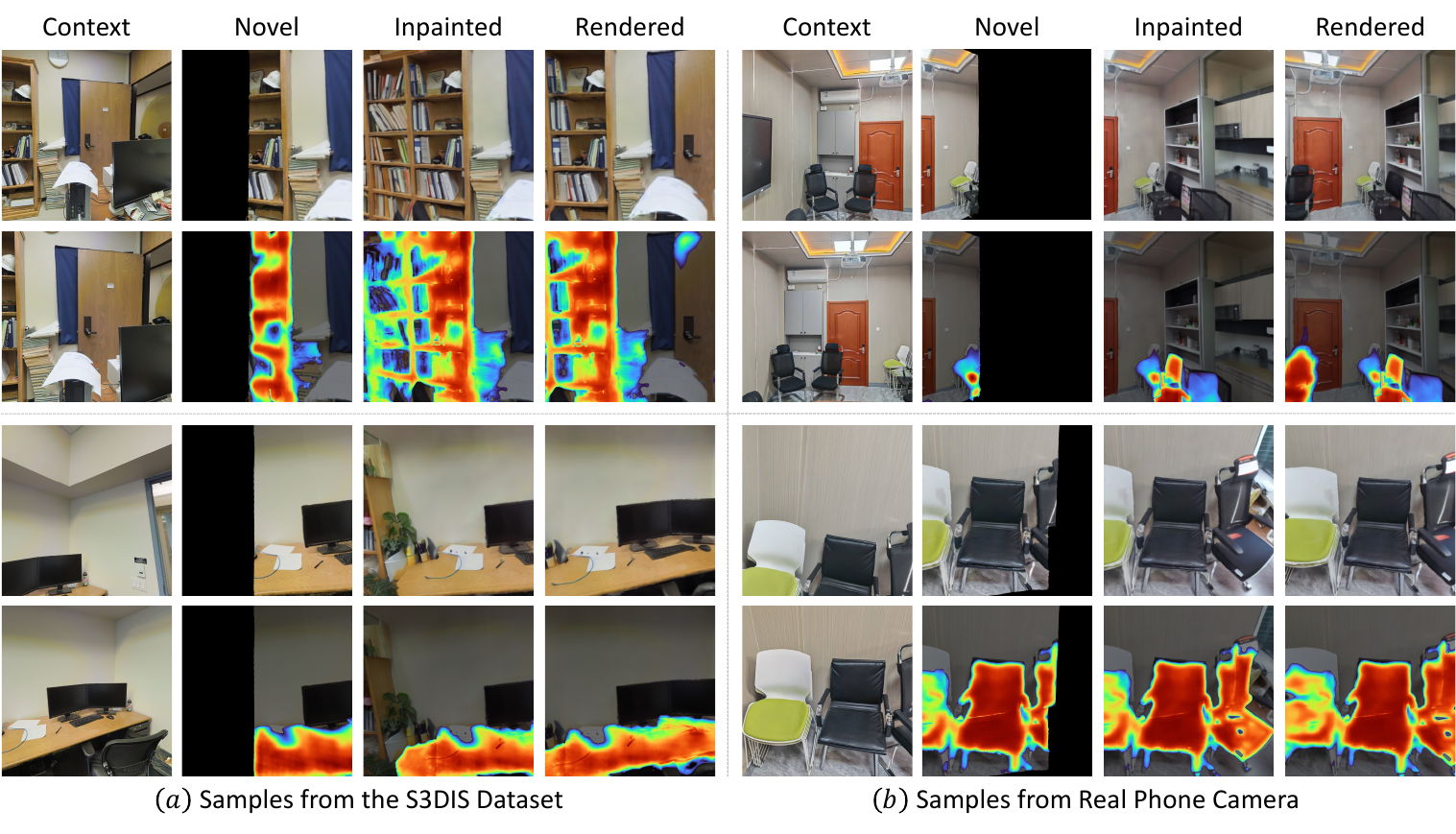}
  \vspace{-18pt}
  \caption{
    \textbf{Model generalization ability evaluation.} Column $(a)$ shows results where the context views are taken from the S3DIS~\cite{armeni2016s3dis}. We query \textit{bookshelf} and \textit{table} for each sample, respectively. In column $(b)$, the context views are captured directly using a \textbf{phone camera}, and we query \textit{chair}.
  }
  \label{fig:more_results}
  \vspace{-12pt}
\end{figure*}

\subsection{Main Results}
\noindent\textbf{Baseline Methods for Comparison.}
We select two representative baselines for comparison: LangSplat~\cite{qin2024langsplat}, a per-scene optimization model, and Splatt3R~\cite{smart2024splatt3r}, a generalizable model. LangSplat relies heavily on accurate initialization via COLMAP~\cite{schonberger2016colmap}, which becomes unreliable when only two input images are available. To address this limitation and enable fair comparison, we initialize LangSplat using point cloud positions predicted by Splatt3R, allowing the model to focus more effectively on learning semantic representations. Meanwhile, as Splatt3R does not support open-vocabulary semantic prediction in its original form, we extend it with a semantic head trained in our first stage in Section~\ref{sec:semantics}. During evaluation, for all models, we consider only the regions rendered by Gaussians with an accumulated opacity greater than 0.01 as valid predictions for computing the IoU scores. This threshold filters out low-confidence regions and ensures consistency across models.

\noindent\textbf{Quantitative Comparisons.}
In Table~\ref{tab:go_benchmark_results}, we compare LangSplat~\cite{qin2024langsplat}, Splatt3R~\cite{smart2024splatt3r}, and OGGSplat on the GO benchmark. OGGSplat consistently outperforms the baselines by a significant margin on both visual fidelity (FID) and semantic plausibility (mIoU). It's worth noticing that the overall FID remains relatively high across all methods. The main reason is the limited number of context pairs available in the validation set, which constrains data diversity. We are unable to sample more pairs because some scenes in the ScanNet++ validation set are relatively small. To maintain a consistent sampling ratio across all validation scenes, we limit the number of context pairs to 10 per scene. 
Regarding semantic plausibility, OGGSplat achieves notably better performance on common large objects such as \textit{chair}, \textit{table}, and \textit{bed}. However, the model performs relatively worse on the \textit{ceiling} class. We attribute this to the limitations of the APE encoding, as well as the difficulty of the Splatt3R backbone in distinguishing between the \textit{ceiling} and \textit{wall} with similar appearance in color and texture. We believe this limitation can be addressed in future work by leveraging more powerful vision-language models and more superior generalizable Gaussian reconstruction methods.

\noindent\textbf{Qualitative Comparisons.}
We conduct extensive qualitative comparisons with baseline methods and illustrate them in Figure~\ref{fig:qualitative_results}. OGGSplat performs better in both novel rendered images and open-vocabulary querying. Regarding rendered images, LangSplat tends to overfit the context views, resulting in blurry renderings from novel viewpoints, even when the Gaussian positions have been initialized. Splatt3R, on the other hand, exhibits large black regions in areas outside the input views. In contrast, OGGSplat reasonably extrapolates unseen regions by leveraging semantic information. Regarding open-vocabulary querying, both LangSplat and Splatt3R are limited to input vision cones. OGGSplat, however, is capable of accurately identifying and querying objects even in previously unseen regions, demonstrating stronger generalization and semantic understanding capabilities.

\noindent\textbf{Model Generalization Ability.}
Apart from ScanNet++ used for training, we also test OGGSplat's generalization ability on data with different distributions. As shown in Figure~\ref{fig:more_results}, OGGSplat successfully reconstructs semantic-aware scenes with an expanded field-of-view using S3DIS~\cite{armeni2016s3dis} samples. We further demonstrate the practicality of OGGSplat on portable devices in column $(b)$, where the context views are captured by a phone camera. The inpainted image and semantic query on \textit{chair} show promising results, highlighting OGGSplat’s potential for applications in daily life.

\begin{figure*}[t]
  \centering
   \includegraphics[width=\linewidth]{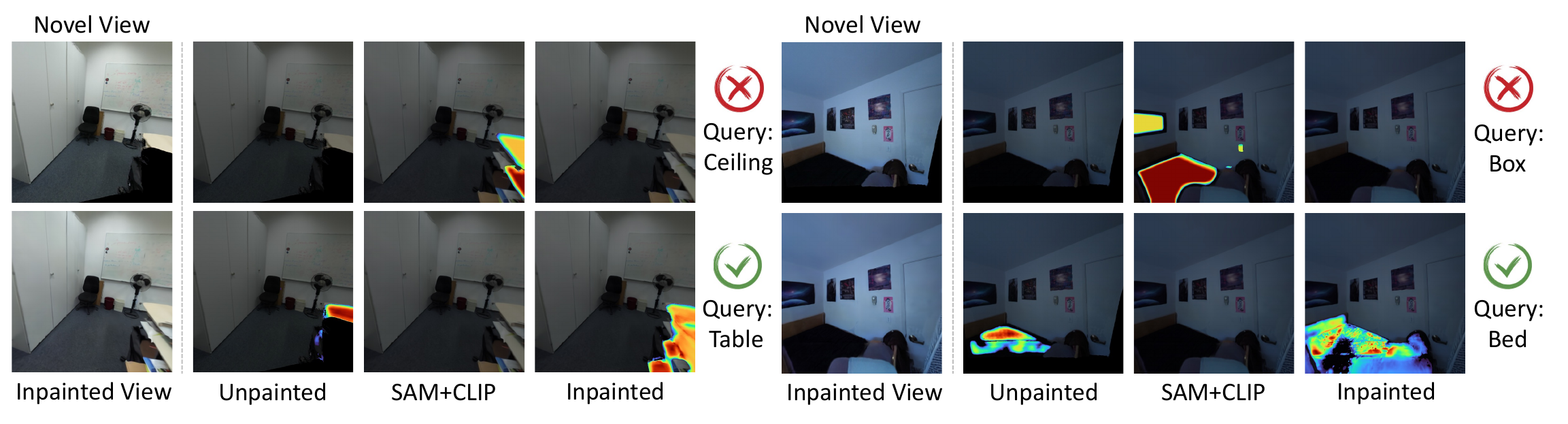}
   \vspace{-18pt}
    \caption{\textbf{Ablations on the effect of semantic diffusion model.} We compare open-vocabulary predictions between the SAM+CLIP offline method and our semantic diffusion inpainting module.}
  \label{fig:ablation_sam}
  \vspace{-10pt}
\end{figure*}

\begin{table}[t]
  \caption{\textbf{Ablations on the GO Benchmark evaluating the impact of the bidirectional control strategy.} The performance is measured by mIoU (\%) across various semantic categories.}
  \vspace{4pt}
  \label{tab:ablation}
  \centering
  \setlength\tabcolsep{5pt}
  \resizebox{\linewidth}{!}{
  \begin{tabular}{ccccccccccccc}
    \toprule
    \multicolumn{2}{c}{Control Type} & \multicolumn{11}{c}{Segmentation Results (IoU $\uparrow$)} \\
    \cmidrule(lr){1-2} \cmidrule(lr){3-13}
     S$\rightarrow$RGB & RGB$\rightarrow$S & \cellcolor{Gray}mIoU  & wall & ceiling  & floor  & table & door & (s)cabinet & chair & (b)shelf & box  & bed \\
    \midrule
    \xmark & \cmark  &  \cellcolor{Gray}16.6  &\textbf{45.8} & 0.1 & 56.8 & 12.3 & 4.6  & 2.8 & 19.3 & 6.3 & \textbf{3.6}  & 15.1   \\
    \cmark & \xmark &  \cellcolor{Gray}14.4 & 43.0 & 0.1 & 47.6 & 10.3 & 5.0 & 3.5 & 16.6 &2.4 & 2.5 & 12.7  \\
    \cmark & \cmark &   \cellcolor{Gray}\textbf{17.6}  & 45.6 & 0.1 & \textbf{58.3} & \textbf{13.3} & \textbf{5.4} & \textbf{3.7} & \textbf{21.4} & \textbf{7.4} & 3.1 & \textbf{18.0} \\
    \bottomrule
  
  \end{tabular}
  }
  \vspace{-10pt}
\label{tab:ablations}
\end{table}

\subsection{Ablation Studies}

In Section~\ref{sec:inpainting}, we introduced the RGB-semantic consistent inpainting module. In this section, we first highlight the importance of the semantic diffusion branch, followed by comprehensive ablations on the GO benchmark to evaluate the effectiveness of the proposed bidirectional control strategy.

\textbf{Semantic Diffusion Model.}  
To obtain reliable semantics for the inpainted regions, we train a semantic diffusion module. A straightforward alternative would be employing an offline open-vocabulary semantic segmentation model, such as SAM~\cite{kirillov2023SAM}+CLIP~\cite{radford2021CLIP} as LangSplat~\cite{qin2024langsplat}. However, this often leads to semantic inconsistency with the original Gaussian, particularly when the objects are partially visible (see Figure~\ref{fig:ablation_sam}). It tends to produce incorrect results even in regions originally correctly predicted, and these errors can propagate and negatively affect the subsequent Gaussian growing. In contrast, our trained semantic diffusion model preserves the semantic consistency in the unpainted regions and significantly improves the accuracy of the predicted semantics in the inpainted areas by leveraging the semantic priors from the visible context. This ensures that the newly generated content aligns well with the existing scene semantics, leading to better overall reconstruction quality.

\begin{figure*}[t]
  \centering
   \includegraphics[width=\linewidth]{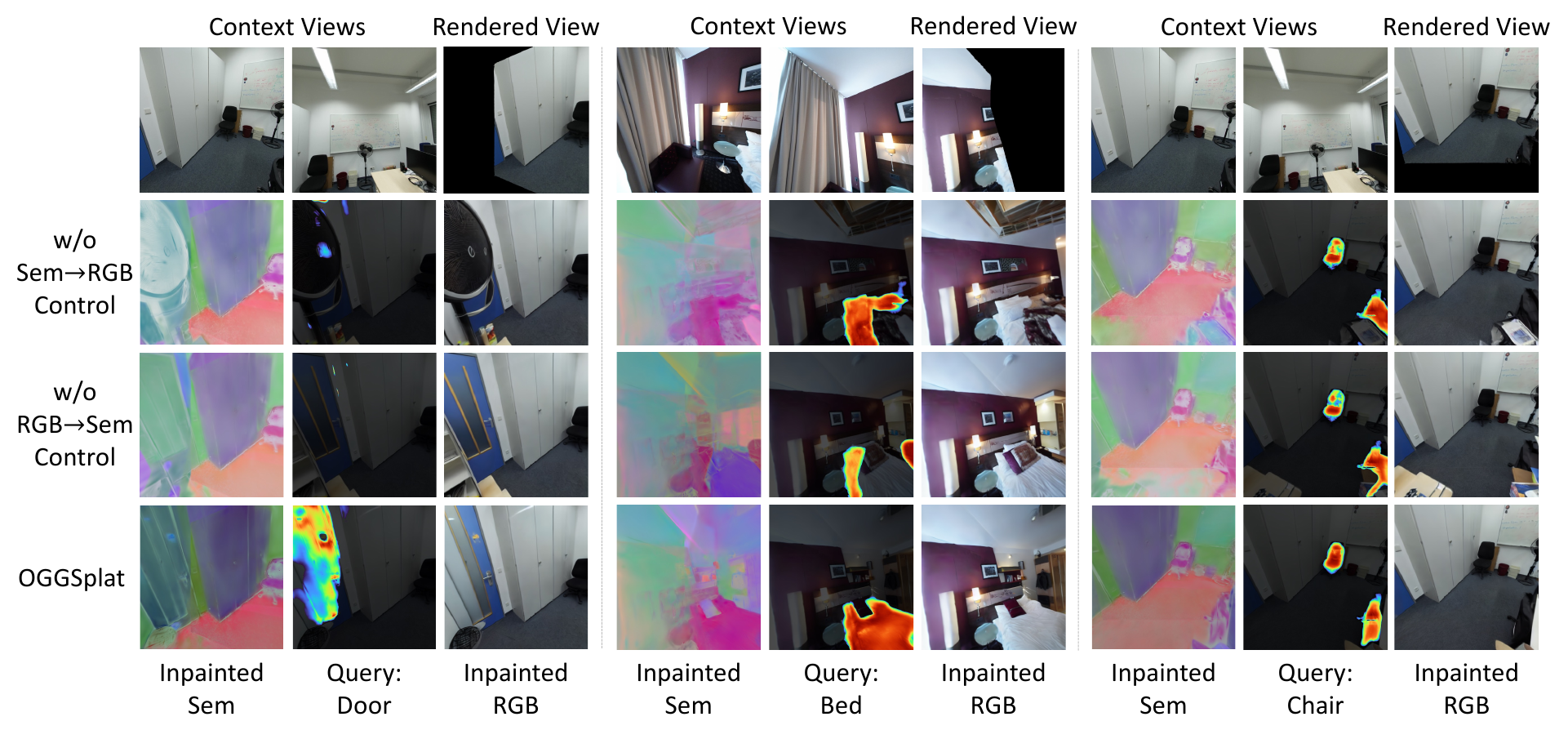}
   \vspace{-18pt}
  \caption{\textbf{Qualitative comparison of bidirectional control.} Row 1 shows the context images and the incomplete renderings from novel views. Rows 2 to 4 correspond to the ablation settings in Table~\ref{tab:ablation}, where each variant removes one of the control mechanisms to examine its individual effect.
  }
  \vspace{-12pt}
  \label{fig:ablation}
\end{figure*}

\textbf{Semantic-to-RGB Control.}  
With access to open-vocabulary semantics, we propose an edge translator to extract semantic cues from the Gaussian boundaries and guide the image/feature completion. In the first row of Table~\ref{tab:ablations}, we remove the edge translator and instead use a generic description (“a room”) as the text prompt. As a result, semantic segmentation performance across most categories decreases. This degradation is also evident in the qualitative comparison in Figure~\ref{fig:ablation}, where the generated content appears more ambiguous and less semantically grounded. These results validate the effectiveness of our semantic-to-RGB control in guiding high-fidelity, semantically consistent Gaussian growth.

\textbf{RGB-to-Semantic Control.}  
In OGGSplat, the semantic inpainting model is explicitly controlled by inpainted images. We remove it in the second row of Table~\ref{tab:ablations} and the third row of Figure~\ref{fig:ablation}. Without RGB-to-semantic control, the generated RGB images and semantic maps exhibit poor spatial alignment, leading to significantly degraded segmentation accuracy. In contrast, introducing the RGB-to-semantic control clearly improves spatial consistency and yields much better performance.

\section{Limitations and Conclusion}
\label{sec:limit}
In this paper, we design OGGSplat, an open Gaussian growing method for generalizable reconstruction with expanded field-of-view. By leveraging semantic cues from open Gaussians and introducing RGB-semantic consistent inpainting via bidirectional controls, our method effectively expands the field-of-view and ensures both visual fidelity and semantic coherence. The reconstructed out-of-view regions are progressively refined through an efficient Gaussian optimization process. To facilitate evaluation, we proposed the Gaussian Outpainting benchmark, which quantitatively assesses the generative and semantic quality of open-vocabulary scene reconstruction. Extensive experiments demonstrate that OGGSplat achieves superior performance in extrapolating beyond the input view cone, marking a significant step forward in generalizable and flexible 3D reconstruction.
However, OGGSplat is currently limited to indoor scenes, since depth estimation in outdoor environments is more challenging, leading to performance decreasing of our baseline model Splatt3R. Nevertheless, we believe that with the integration of more powerful and generalizable Gaussian reconstruction models in the future, our approach can achieve promising performance in outdoor scenarios as well.

\appendix

\section{Additional Experimental Results}
\subsection{Video Results}
To provide a more comprehensive and intuitive visualization of our method, we include video results in the supplementary ZIP file. Specifically, we present visualizations across five different scenes. For each scene, we showcase the rendering results of both Splatt3R~\cite{smart2024splatt3r} and OGGSplat under continuous camera views. Additionally, we provide the corresponding relevance score heatmaps under a specific open-vocabulary query, enabling a direct comparison of semantic understanding across the two methods. As clearly demonstrated, our model effectively extrapolates to unseen regions while maintaining both high visual fidelity and semantic plausibility.
\subsection{Ablation on Separate Diffusion UNet}

\begin{figure*}
  \centering
  \includegraphics[width=\linewidth]{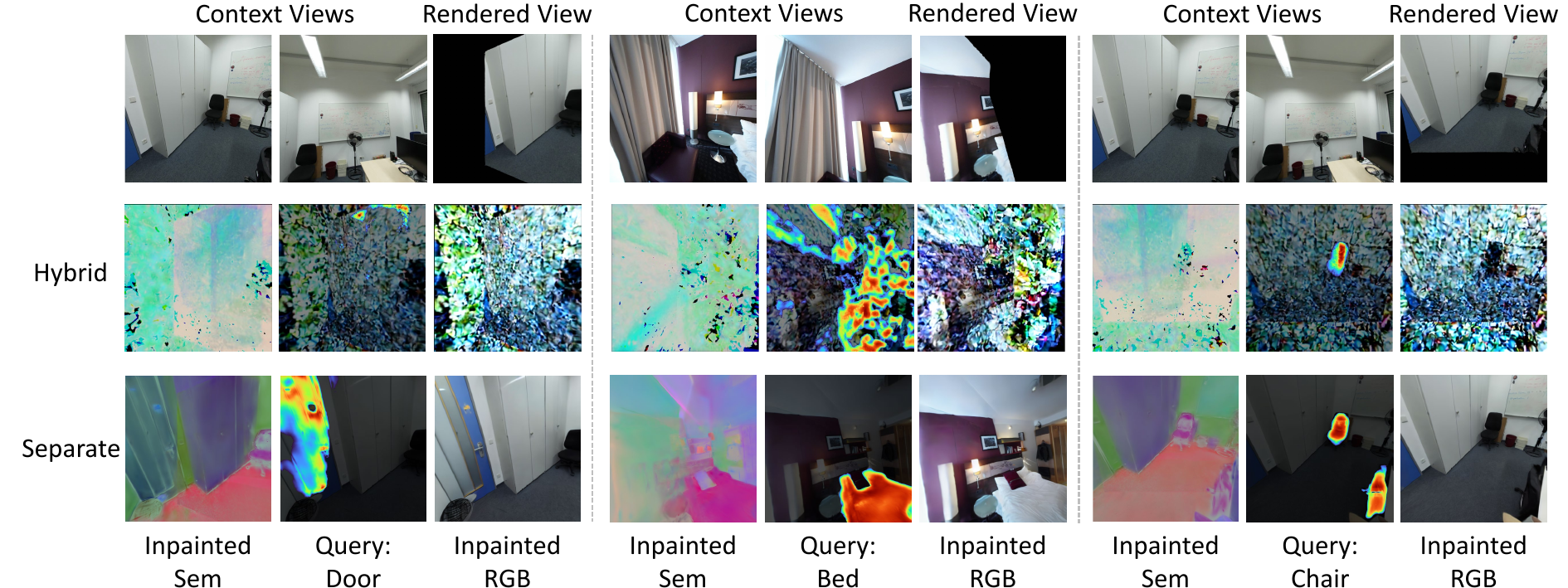}
  \vspace{-8pt}
  \caption{ Qualitative comparison between hybrid (shared-weight) and separate diffusion UNet architectures. Row 1 shows the context images along with the incomplete renderings from novel views. Row 2 presents the results by using a hybrid UNet that jointly predicts RGB image and semantic content using shared weights. Row 3 shows the results from our proposed architecture with two separate UNets: one for RGB image synthesis and the other for semantic prediction. 
  }
  \label{fig:supp_ablation}
  \vspace{-12pt}
\end{figure*}

To enable the generation of both spatially consistent RGB images and semantic content, we train two separate diffusion models: $\mathrm{Diff}_{\mathrm{rgb}}$ and $\mathrm{Diff}_{\mathrm{sem}}$, and enforce spatial consistency between them using a ControlNet~\cite{zhang2023controlnet}-based approach. 
A simpler alternative would be to employ a single shared diffusion UNet based on an image diffusion model~\cite{rombach2022ldm}, modified to allow additional semantic inputs and outputs by adjusting the input and output convolutional channels. However, our experiments show that this approach fails to produce meaningful RGB and semantic outputs. As illustrated in Fig.~\ref{fig:supp_ablation}, using a hybrid (shared) diffusion UNet leads to severe distortions in both RGB images and semantic content. 
We think that this failure is due to the significant differences between the latent spaces of the RGB image VAE and the semantic VAE, which makes it difficult for a single UNet to learn consistent mappings in both domains. These results highlight the effectiveness and necessity of our separate $\mathrm{Diff}_{\mathrm{sem}}$ model and the corresponding control module design.

\section{Implementation Details}
\subsection{Scannet++ Dataset}
ScanNet++ dataset~\cite{yeshwanth2023scannet++} provides high-quality 3D geometry along with high-resolution RGB images of various indoor environments. Following the protocol introduced by Splatt3R, originally designed for 3D reconstruction, we adopt the standard training split comprising 230 scenes and the validation split containing 50 scenes. Following~\cite{smart2024splatt3r,wang2024gsemsplat}, we also discard frames missing reliable depth information. All selected frames are uniformly cropped and resized to a spatial resolution of $512 \times 512$.

\subsection{Training Settings }

\label{subsec: traing_settings}

To provide a clearer overview of the experimental configurations used at different training stages, we summarize the details in Table~\ref{tab:supp_settings}. The table includes the settings for all key components that need to be trained in our method, namely the generalizable open Gaussian initialization module, RGB UNet, semantic VAE~\cite{kingma2013vae}, semantic UNet, ControlNet~\cite{zhang2023controlnet}, and the open Gaussian growing process. 
\noindent\textbf{Generalizable Open Gaussian Initialization}. We adopt the pretrained Splatt3R model and freeze its backbone, which is responsible for predicting the basic Gaussian attributes. We then train only the newly added semantic head, denoted as \( H_{\text{sem}} \). During training, we use two context images as input and supervise the model by rendering three target views from the training split. 
Following the setup in Splatt3R~\cite{smart2024splatt3r}, the context images are selected such that at least 30\% of the pixels in the second image have direct correspondences in the first image. Similarly, target images are chosen such that at least 30\% of their content is visible in at least one of the context images.

\noindent\textbf{RGB-Semantic Consistent Inpaintor}. For RGB image inpainting model $\mathrm{Diff}_{\mathrm{sem}}$, we fine-tune a stable diffusion inpainting model~\cite{rombach2022ldm} to better align the generated appearance with realistic indoor scenes. 
In addition to standard RGB inpainting, we propose a novel diffusion-based feature inpainting model, denoted as $\mathrm{Diff}_{\mathrm{sem}}$, which consists of both a Variational Autoencoder~\cite{kingma2013vae} (VAE)  and a UNet architecture. This model enables semantic-aware inpainting in the feature space while maintaining consistency with the RGB domain.
To ensure spatial consistency between the RGB and semantic contents, we train an auxiliary RGB control module inspired by ControlNet~\cite{zhang2023controlnet} that guides the inpainting process in the feature space.

\begin{wraptable}{r}{0.42\textwidth}
\vspace{-10pt}

\captionsetup{width=0.42\textwidth}
\centering

\caption{Learning rates for different Gaussian parameters.}
\label{tab:supp_gaussian_settings}
\begin{tabular}{@{}lc@{}}
\toprule
\textbf{Parameter} & \textbf{Learning Rate} \\
\midrule
point position $\mu$ & 1e-2 \\
rotation quaternion $q$ & 1e-3 \\
scale vector $s$ & 5e-3 \\
opacity scalar $\alpha$ & 5e-2 \\
spherical harmonics $\mathbf{S}$ & 2.5e-2 \\
semantic feature $f$ & 2.5e-3 \\
\bottomrule
\end{tabular}
\vspace{-10pt}
\end{wraptable}
\noindent\textbf{Open Gaussian Growing}.
We set the horizontal and vertical outpainting angles to lie within the ranges of $[-60^\circ, 60^\circ]$ and $[-20^\circ, 20^\circ]$, respectively. To simplify this stage, we decouple the horizontal and vertical rotations: when the horizontal angle is non-zero, the vertical angle is set to zero, and vice versa. For each optimization round, to improve efficiency, we use two inpainted images and their corresponding semantic maps under symmetrical camera poses to provide the supervision signal. Moreover, the selected camera view pairs are arranged to exhibit progressively increasing angular differences, thereby enabling a gradual and progressive Gaussian growing process. Specifically, denoting the camera rotation angles in the horizontal and vertical directions as $(\theta_h, \theta_v)$, the sampled camera angles are selected in the following order:
$(0^\circ, 0^\circ)$, 
$(0^\circ, \pm20^\circ)$, 
$(\pm30^\circ, 0^\circ)$, and 
$(\pm60^\circ, 0^\circ)$. It is worth noting that during actual optimization, camera poses can be arbitrary. This sampling strategy is adopted purely to facilitate a simpler, more consistent, and computationally efficient optimization process.
We conduct a total of four optimization rounds. In the first round, we perform inpainting without changing the camera poses, i.e., using poses of the original context views. This step focuses on refining low-confidence regions through inpainting to enhance rendering quality under the original views. In subsequent rounds, we fix the batch size to 4 and include supervision signals from the originally inpainted context views, previously inpainted views, and newly generated inpainted views. For the optimization of Gaussian parameters, we adopt parameter-specific learning rates following the setting proposed in~\cite{qin2024langsplat}. The detailed learning rates for each type of parameter are summarized in Table~\ref{tab:supp_gaussian_settings}. Empirically, we observe that each optimization round converges efficiently within 600 training iterations.

\begin{table}[t]
\setlength\tabcolsep{5pt}
  
  \caption{Experiment settings for different training stages.}
  \label{tab:supp_settings}
  \vspace{2pt}
  \newcolumntype{g}{>{\columncolor{Gray}}c}
  \small
  \setlength\tabcolsep{2pt}
    \begin{tabular}{l|c|c|c|c|c|c}
    \toprule
    \multirow{2}{*}{Config}&\multirow{2}{*}{Gaussian Init.} &\multicolumn{4}{c|}{RGB-Semantic Consistent Inpaintor} &  \multirow{2}{*}{Gaussian Growing}
    \\
      &  &  RGB UNet &  Sem. VAE & Sem. UNet &  ControlNet & \\
    \midrule
    optimizer       & Adam & AdamW8bit & AdamW & AdamW8bit  & AdamW8bit & Adam\\
    learning rate   & 1e-5 & 1e-5 & 6e-6 & 1e-5  & 1e-5 & hybrid (Table~\ref{tab:supp_gaussian_settings})\\
    weight decay    & 5e-2 & 1e-2 & 1e-2 & 1e-2 & 1e-2 & 0\\ 
    scheduler & multi-step  & constant & cosine & constant  & constant & exponential\\
    batch size      & 12 & 4 & 2 & 4  & 4 & 4\\
    accumulation steps & 1 & 2 & 4 & 2  & 2 & 1\\
    training iterations & 500,000 & 50,000 & 45,000 & 20,000  &10,000 & 600\\
    GPU device      & 8 RTX 3090 & 8 RTX 3090 & 8 RTX 3090 & 8 RTX 3090 & 8 RTX 3090 & 1 RTX 3090\\
    image size      & 512$\times$512 & 512$\times$512 & 512$\times$512 & 512$\times$512 & 512$\times$512 & 512$\times$512\\

    \bottomrule
  \end{tabular}
  \vspace{-10pt}
\end{table}
\subsection{RGB-to-Semantic ControlNet Module}

\setlength{\columnsep}{10pt}
\setlength{\intextsep}{-10pt}
\begin{wrapfigure}{r}{7cm} \small
\centering
\includegraphics[width=\linewidth]{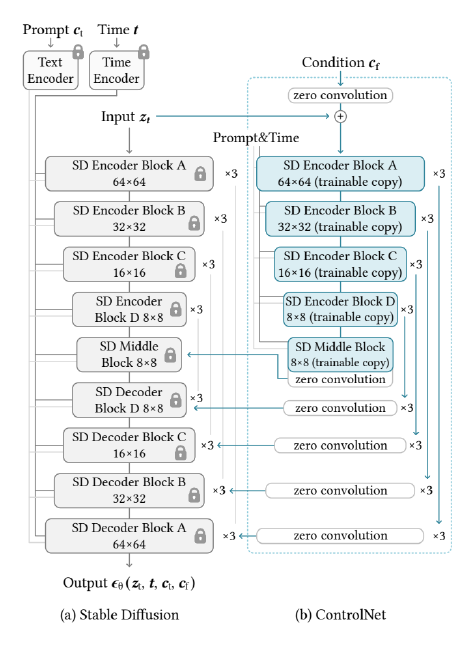}
\setlength{\abovecaptionskip}{-20pt}
\caption{\small The architecture of the ControlNet~\cite{zhang2023controlnet}.}
\label{fig:supp_controlnet}
\end{wrapfigure}

To ensure spatial alignment between the inpainted RGB image and its corresponding semantic map, we adopt a control mechanism inspired by ControlNet~\cite{zhang2023controlnet}, where the RGB image serves as guidance for the generation of the semantic map. An overview of the ControlNet architecture is illustrated in Fig.~\ref{fig:supp_controlnet}. Specifically, our control module comprises the encoder and bottleneck components of the stable diffusion UNet architecture, with their weights initialized from the corresponding layers of a pretrained stable diffusion UNet. Conditional signals are then injected into the bottleneck and decoder parts via zero convolutions and element-wise addition. To accelerate training and enhance the effectiveness of control learning, we initialize the control module with pretrained parameters from a ControlNet model~\cite{zhang2023controlnet} conditioned on image segmentation. This initialization strategy provides a strong prior for spatially consistent generation and significantly improves both training efficiency and overall performance. Details of the training settings for this module are provided in Table~\ref{tab:supp_settings}.

\subsection{GO Benchmark}
\label{subsec: gO_benchmark}
For evaluation on our proposed GO Benchmark, we uniformly sample \textit{16 novel camera poses} around the context image $I_1$, covering a horizontal angular range of $[-60^\circ, 60^\circ]$ and a vertical angular range of $[-20^\circ, 20^\circ]$. To simplify the evaluation setup, we decouple horizontal and vertical rotations, following the same strategy described in Section~\ref{subsec: traing_settings}.
The IoU score for every query is computed by averaging over a total of $50 \times 16$ images. If the union of predicted and ground-truth regions in an image is empty, that image is excluded from the IoU computation. 
To ensure robustness, we repeat the inpainting, growing, and evaluation process five times with the same settings and report the mean IoU as the final benchmark result.

\subsection{Relevance Score for Evaluation}

During open-vocabulary querying, we select regions with a relevance score greater than 0.5 as the final predicted category mask. The computation of the relevance score is inspired by prior works~\cite{kerr2023lerf,qin2024langsplat,shi2024legaussian}, and is defined as follows for each query:

\begin{equation}
\text{Relevance} = \min_i \frac{\exp(g_{\text{img}} \cdot g_{\text{qry}})}{\exp(g_{\text{img}} \cdot g_{\text{qry}}) + \exp(g_{\text{img}} \cdot g_{\text{canon}}^i)},
\end{equation}

where \(g_{\text{img}}\) denotes the image semantic feature, \(g_{\text{qry}}\) is the query APE embedding, and \(g_{\text{canon}}^i\) represents the APE embedding of a predefined canonical phrase such as \textit{"object"}, \textit{"things"}, \textit{"stuff"}, or \textit{"texture"}.

In contrast to the mentioned prior works, which typically focus on a limited set of categories in a single scene and require the set of possible scene categories to be known in advance, we adopt a more general strategy. These prior methods often normalize the relevance score and select masks based on a threshold over the normalized values. However, this approach may incorrectly force the prediction of masks even for categories absent in the scene. To address this limitation and enhance generalizability, we directly apply a fixed threshold of 0.5 to the raw (unnormalized) relevance scores and select pixels with scores exceeding this threshold as the final predicted mask. This ensures that only queries with truly high relevance scores produce predictions, avoiding false positives in irrelevant categories. As a result, we are able to compute per-category prediction masks from a predefined query set without requiring manual query specification for each individual scene.

\bibliographystyle{plain}
\bibliography{ref}

\end{document}